\documentclass[10pt,twocolumn,letterpaper]{article}

\usepackage[pagenumbers]{cvpr} 
\usepackage{graphicx}
\usepackage{amsmath}
\usepackage{amssymb}
\usepackage{booktabs}
\usepackage{multirow}
\usepackage{subfloat}
\usepackage{algorithm}
\usepackage{algorithmic}
\usepackage[toc,page]{appendix}
\usepackage[vlined,algo2e,linesnumbered]{algorithm2e}
\SetCommentSty{mycommfont}
\SetKwInput{KwInput}{Input}                
\SetKwInput{KwOutput}{Output}              
\SetKwProg{Fn}{Function}{:}{}
\newcommand{\tabincell}[2]{\begin{tabular}{@{}#1@{}}#2\end{tabular}}
\usepackage{pifont}
\newcommand{\cmark}{\ding{51}}%
\newcommand{\xmark}{\ding{55}}%
\def\dotfill#1{\cleaders\hbox to #1{·}\hfill}
\newcommand\dotline[2][.5em]{\leavevmode\hbox to #2{\dotfill{#1}\hfil}}
\newcommand*{\affaddr}[1]{#1} 

\newcommand*{\email}[1]{\texttt{#1}}

\usepackage[pagebackref,breaklinks,colorlinks]{hyperref}

\usepackage[capitalize]{cleveref}
\crefname{section}{Sec.}{Secs.}
\Crefname{section}{Section}{Sections}
\Crefname{table}{Table}{Tables}
\crefname{table}{Tab.}{Tabs.}

\begin{document}

\title{NomMer: Nominate Synergistic Context in Vision Transformer  \\ for Visual Recognition}

\author{%
	Hao Liu\textsuperscript{\dag\thanks{Equal contribution. \textsuperscript{\dag}Contact person.}} \quad Xinghua Jiang\textsuperscript{*}  \quad Xin Li \quad Zhimin Bao \quad Deqiang Jiang \quad Bo Ren\\		
	\affaddr{Tencent YouTu Lab} \quad\\
	\email{\small \{ivanhliu, clarkjiang, fujikoli, zhiminbao, dqiangjiang, timren\}@tencent.com}
}
\maketitle
\begin{abstract}
 Recently, Vision Transformers~(ViT), with the self-attention~(SA) as the de facto ingredients, have demonstrated great potential in the computer vision community. For the sake of trade-off between efficiency and performance, a group of works merely perform SA operation within local patches, whereas the global contextual information is abandoned, which would be indispensable for visual recognition tasks. To solve the issue, the subsequent global-local ViTs take a stab at marrying local SA with global one in parallel or alternative way in the model. Nevertheless, the exhaustively combined local and global context may exist redundancy for various visual data, and the receptive field within each layer is fixed. Alternatively, a more graceful way is that global and local context can adaptively contribute per se to accommodate different visual data. To achieve this goal, we in this paper propose a novel ViT architecture, termed NomMer, which can dynamically \textbf{Nom}inate the synergistic global-local context in vision transfor\textbf{Mer}. By investigating the working pattern of NomMer, we further explore what context information is focused. Beneficial from this ``dynamic nomination'' mechanism, without bells and whistles, the NomMer can not only achieve 84.5\% Top-1 classification accuracy on ImageNet with only 73M parameters, but also show promising performance on dense prediction tasks, i.e., object detection and semantic segmentation. The code and models are publicly available at~\url{https://github.com/TencentYoutuResearch/VisualRecognition-NomMer}.

\end{abstract}

\vspace{-2mm}
\section{Introduction}
\label{sec:intro}

\maketitle
\begin{figure}[htb]
	\includegraphics[width=1\linewidth]{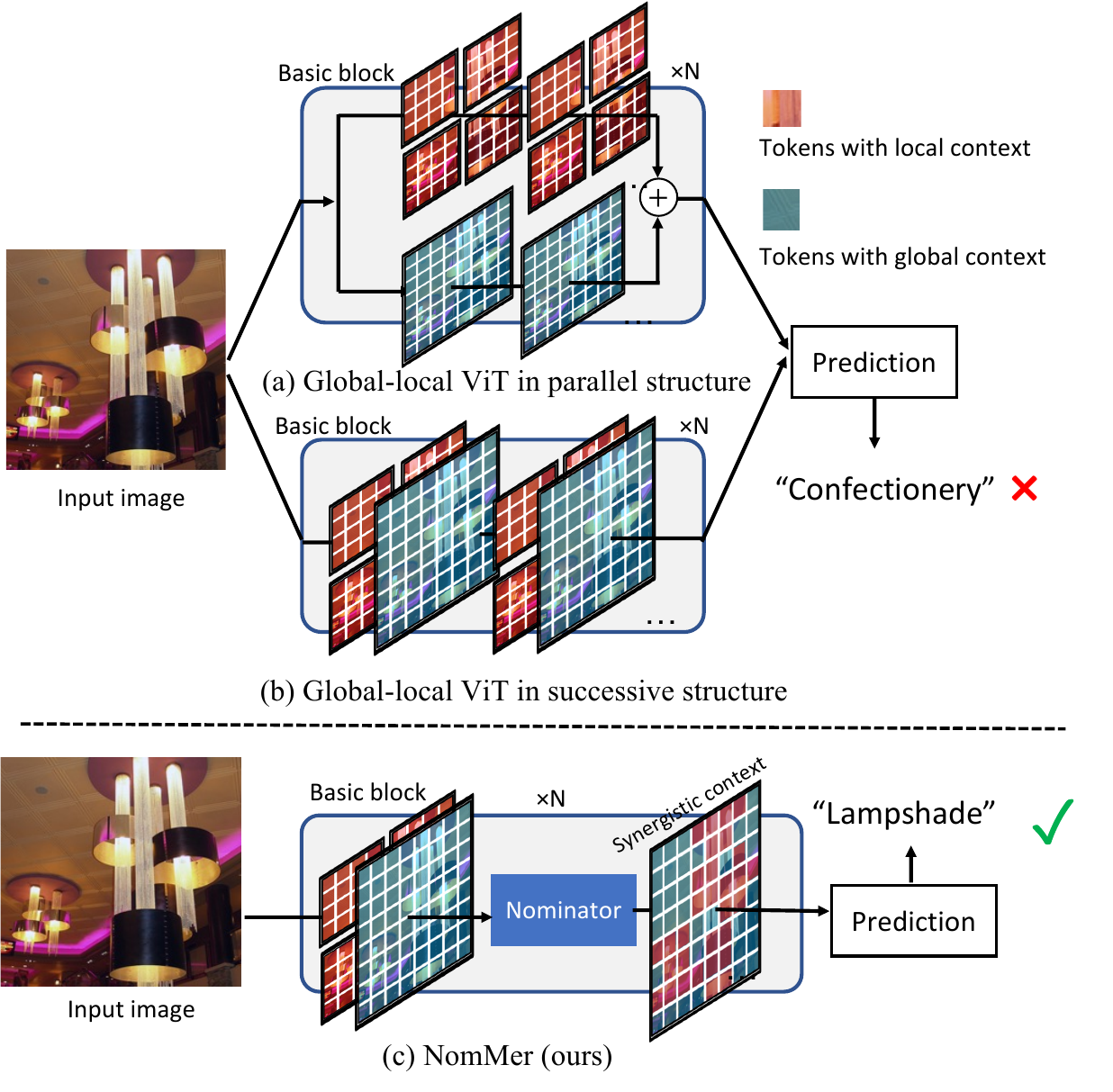}
	\vspace{-2.5mm}
	\caption{Illustration of motivation of the proposed NomMer. (a)~Global-local ViTs in parallel structure. (b)~Global-local ViTs in successive structure. Previous global-local ViTs only focus on fusing global and local contexts while the modulation on them lacks, where the redundant information may have negative impact when recognizing various cases. (c) Our NomMer. When recognizing an object, our method can dynamically yield synergistic context from global-local context through nominator.} 
	\label{fig:moti}
\end{figure}

In computer vision, Convolutional Neural Networks~(CNNs)~\cite{he2016deep, chen2019drop, krizhevsky2012imagenet, simonyan2014very, szegedy2017inception, xie2017aggregated} have served as
the \textit{de facto} gold standard for many years. Recently, the Vision Transformer~(ViT)~\cite{dosovitskiy2020image} and its variants~\cite{wang2021pyramid, touvron2021training} have been proposed challenging the \textit{status quo}. Thanks to the global information communication and content-dependent learning nature of Self-Attention~(SA), which is substantively different from  local behavior in CNN, ViTs have showed superior performance on many visual recognition tasks.

However, the ViTs reasoning global dependency of split feature patch embeddings~(\textit{a.k.a}, tokens) is computationally expensive. To attack the issues, many recent works, such as SwinT~\cite{liu2021swin}, TNT-ViT~\cite{yuan2021tokens} and HaloNet~\cite{vaswani2021scaling}, introduce CNN-like inductive bias~(\textit{e.g.}, locality, translation equivariance) by building the token relations via self-attentions only within local windows and hierarchically aggregating them in a bottom-up manner. These local SA-based ViTs significantly improve the data efficiency, whereas the global contextual relation, especially in the early stage, is abandoned. To remedy this shortage, as illustrated in Fig.~\ref{fig:moti}~(a) and (b), two kinds of ``global-local ViTs'' take a compromise, where both local and global visual dependencies are incorporated for each token. The first kind~\cite{li2021local,yang2021focal, peng2021conformer} 
marries the context aggregated by local SA or CNN with the one captured by global SA in parallel structure, such as multi-granular connection~\cite{szegedy2015going}, in the basic block of hierarchical ViT. On the other hand, the local and global context are aggregated alternatively in the second kind~\cite{zhang2021aggregating, wu2021cvt}. However, no matter in the parallel or successive manner, these global-local ViTs only focus on fusing global-local context while the modulation on it lacks, where the redundant information may have negative impact when recognizing various cases. For example, the ``lampshade''  is mis-recognized as ``confectionery'' by previous method, probably due to redundant contextual clues, such as colored lights, misleading the model.

According to a pioneering research~\cite{stewart2020review}, the human visual system can simultaneously process both \textit{peripheral} and \textit{foveal} vision when perceiving the real world scene, which also exhibits the intriguing property of modulation under various scenes. As a result, the redundant information can be naturally ignored. Specifically, foveal vision focuses on a local region of interest with more visual details, such as fine-grained details, colors or textures of objects in the scene, while peripheral vision refers to the one viewed at large angles but containing rough global scene information. 

Inspired by the above preliminary and observations, in this paper, we regard the locally aggregated context in global-local ViT model as   
the foveal vision, while the globally aggregated one is treated as the peripheral-like visual information. Moreover, we propose a more effective context leverage strategy that the useful dependency information is nominated from local and global context dynamically. To achieve this nomination process, we need to solve the following two non-trivial problems: (i)~\textit{How to make nominated global and local context work in harmony when processing different visual cases?} (ii)~\textit{How to preserve the information at most without increasing computational cost evidently when reasoning global dependency?}  

For the first problem, a straightforward way of nomination is to directly hard-sample from global or local context. Unfortunately, this sample process is indifferentiable, which makes the model suffer from the gradient lost problem. To overcome this issue, we coin a novel learnable Synergistic Context Nominator~(SCN) to dynamically yield the global-local context with synergy for each spatial location, which is vividly shown by Fig.~\ref{fig:moti}~(c). Additionally, as for the second global context reasoning problem, most of previous methods~\cite{peng2021conformer, li2021local} adopt the pooling or bilinear down-sampling before conducting global SA-based context aggregation to strike favorable efficiency/performance trade-offs. Nevertheless, this naive computational simplification may remove both redundant and salient information, leading to the detriment on performance. Considering there exists tremendous redundancy in natural images containing most smooth signals with high frequency noise, we build a Compressed Global Context Aggregator~(CGCA) upon Discrete Cosine Transform~(DCT)~\cite{ahmed1974discrete} to reason the global dependency with redundancy reduced from the frequency domain but without increasing computational complexity. This global context reasoning mechanism is also surprisingly consistent with the working behavior of human visual system~\cite{stewart2020review} when processing peripheral vision. 

Based on the SCN and CGCA submodules, we carve out the basic Transformer blocks and stack them into our NomMer framework, which can dynamically \textbf{Nom}inate the synergistic context in vision transfor\textbf{Mer} for various visual data. We have experimentally verified the effectiveness of our proposed method on image classification task as well as dense prediction tasks,~\textit{i.e.}, object detection and semantic segmentation. Our contributions can be summarized as follows:

\noindent 1) We propose a novel learnable Synergistic Context Nominator~(SCN) in terms of aggregated context, which is in stark contrast to previous ViTs with global-local context greedly aggregated.

\noindent 2) We coin a novel Compressed Global Context Aggregator~(CGCA) more effective to reduce global redundancy and to capture global correlations.

\noindent 3) We propose a novel ViT framework, termed NomMer, which enables the nominated global-local context complementary with each other for various cases and tasks. We also investigate the working behavior of SCN in NomMer.

\noindent 4)  Thanks to the ``nomination'' mechanism, the NomMer can achieve 84.5\% Top-1 classification accuracy on ImageNet with only 73M parameters. With fewer parameters, our small and tiny models can still achieve 83.7\% and 82.6\% accuracy respectively.  We also witness the promising performance of NomMer on dense prediction tasks, \textit{i.e.}, object detection and semantic segmentation. 

\section{Related Work}
\label{sec:rw}
\subsection{Vision Transformer}
\noindent \textbf{Global Vision Transformer.}\quad
The vanilla Vision Transformer~(ViT)~\cite{dosovitskiy2020image} greedily reasons the global dependency of visual tokens throughout the whole model. However, it may suffer from the slow model convergence and expensive computational cost. To alleviate it, DeiT~\cite{touvron2021training} exploits distillation technique while PVT~\cite{wang2021pyramid} introduces the pyramid structure~\cite{lin2017feature} into ViT.

\noindent \textbf{Local Vision Transformer.} \quad Due to the lacking of inductive bias and the high computational complexity of SA calculated in global range, the inherent shortages of global ViT cannot be totally eliminated. Therefore, many subsequent ViTs alternatively propose to limit the token relation building by Self-Attention~(SA) only within local regions. Among, SwinT~\cite{liu2021swin} coins basic block with the successive WMHSA and shifted ones to perform within- and cross-window information communication. TNT-ViT~\cite{yuan2021tokens} proposes to represent the local structure by recursively aggregating neighboring Tokens into one Token. In the similar spirit, HaloNet~\cite{vaswani2021scaling} introduces a non-centered local attention and extends it with ``haloing'' operation. Although the local SA-based ViTs significantly reduce the model complexity and improve the data efficiency, whereas the global contextual relation, especially in the early stage, is abandoned. 

\noindent \textbf{Global-Local Vision Transformer.} To overcome the ``global contextual information lost'' issue, many global-local ViTs~\cite{li2021local,yang2021focal, peng2021conformer, zhang2021aggregating, wu2021cvt} attempt to seek an equilibrium between the \textit{fully global} and \textit{fully local} contextual information leverage.   
Specifically, the literature~\cite{yang2021focal} proposes a focal SA incorporates both fine-grained local and coarse-grained global interactions. Conformer~\cite{peng2021conformer} fuses local features and global representations under different resolutions based on the Feature Coupling Unit. Other than the above global-local ViTs with parallel structure, NesT~\cite{zhang2021aggregating} stacks the local SA and CNN-bsed global aggregation modules alternatively while the basic block consisting of successive global SA and local CNN are designed in the CVT~\cite{wu2021cvt}.
\subsection{Redundancy Reduction Methods}
The design of local~\cite{yuan2021tokens, vaswani2021scaling} and global-local ViTs~\cite{zhang2021aggregating} can be regarded as the redundancy reduction in terms of architecture. By contrast, our proposed architecture can dynamically harmonize the local and global context through nomination mechanism. Moreover, OctConv~\cite{chen2019drop} is a pioneering successful attempt on feature redundancy reduction via separating the CNN feature into low- and high-frequencies, which is realized by down-sampling and up-sampling. DRConv~\cite{chen2021dynamic} and DynamicViT~\cite{rao2021dynamicvit} are another two representative works. The DRConv proposes to dynamically select the CNN filters whereas there is still local context exploited, while the DynamicViT dynamically sparsifies tokens, which may underperform on dense prediction tasks due to the attenuation of fine-grained local interactions.
The DCTransformer~\cite{nash2021generating} transits the view of solving the problem into frequency domain and demonstrates the sparse representations can carry sufficient information for generating images. Similarly, the work~\cite{xu2020learning} also converts the input image into frequency domain for visual understanding. The intriguing property of frequency domain motivates our method aggregating global context at feature-level in frequency domain, which is different from above both works only operating input images.    
\begin{figure*}[htb]
	\begin{center}
		\includegraphics[width=1\linewidth]{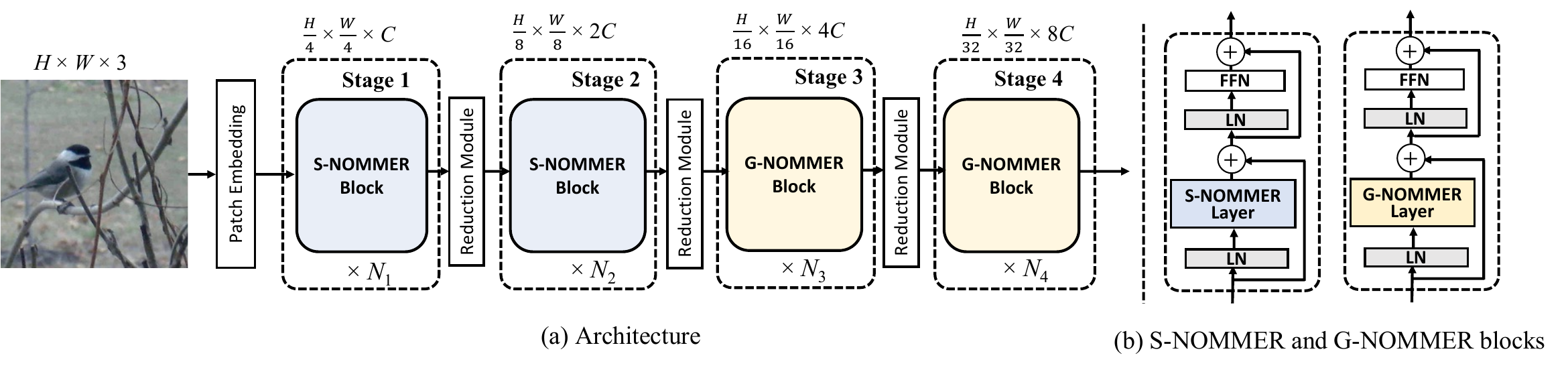}
	\end{center}
	\vspace{-3.5mm}
	\caption{(a)~Architecture of the proposed NomMer; (b)~NomMer basic block consisting of NomMer layer, Layer Normalization~(LN) and Feed-Forward Network(FFN) with residual connections.} 
	\label{fig:arch}
\end{figure*}
\section{Methodology}

\subsection{Architecture Overview}
The overview of the proposed architecture is shown in Fig.~\ref{fig:arch}~(a), which follows the hierarchical design in other ViT models~\cite{liu2021swin,peng2021conformer,yang2021focal,wang2021pyramid}. It mainly consists of four stages with ~``Reduction Modules''~dispersed between adjacent stages to down-sample the token numbers and increase the channels by factor 2. Each reduction module is composed of a $3\times 3$ convolutional layer and a max pooling layer with stride 2. 

Before being sent to the first stage, the image in size of $H\times W \times3$ is split into patches of $4 \times 4 \times 3$ size, which are then projected by ``patch embedding'' consisting of convolutional layer with $4 \times 4 \times C$ kernel and 4 stride size into $\frac{H}{4} \times \frac{W}{4} $ visual tokens in $C$ dimension. We coin two kinds of NomMer basic blocks,\textit{ i.e.}, Synergistic NomMer~(S-NomMer) and Global NomMer~(G-NomMer) blocks in our proposed architecture. Stage 1 and 2 contains $N_1$ and $N_2$  
S-NomMer blocks, which are responsible for capturing the synergistic context from global-local context. As the feature maps in stage 3 and 4 become relatively small in spatial size~($\frac{H}{16} \times \frac{W}{16}$ and $\frac{H}{32} \times \frac{W}{32}$), the global context and local context could become homogeneous and the information is highly abstracted. Therefore, inside the Stage 3 and 4, we build the G-NomMer layers upon Global Self-Attention~(SA), which pay more attention on capturing the semantic information in global range. As is shown in Fig.~\ref{fig:arch}~(b), both S-NomMer and G-NomMer layers are all equipped with Feed-Forward Networks~(FFN)~\cite{vaswani2017attention}, residual connection and Layer Normalization~(LN)~\cite{vaswani2017attention}. As the S-NomMer layer is the core component of our method, we will elaborate it in the next subsections.

\subsection{Synergistic NomMer Layer}

To implement the ``dynamic nomination'' of S-NomMer layer serving as the core ingredient in our proposed ViT model, our design mainly concentrates on two vital aspects: 1) \textit{generating global and local context}; 2) \textit{nominating synergistic context from them for each visual tokens}. Generally, as illustrated in Fig.~\ref{fig:nommer}, S-NomMer layer has three trainable submodules, \textit{i.e.},~Compressed Global Context Aggregator~(CGCA), Local Context Aggregator~(LCA) and Synergistic Context Nominator~(SCN). The CGCA provides rough global context while LCA yields more detailed contextual information within local region. By evaluating the contributions of local or global context aggregated to each tokens, SCN flexibly modulates them and build the nomination map to mask out the features with synergistic context.

\subsubsection{Compressed Global Context Aggregator}
As introduced in Sec.~\ref{sec:intro}, we expect the peripheral-like global context can provide the rough  visual clue in terms of the scene, where more fine-grained information can be supplemented from local context. Therefore, the feature maps including visual tokens for global context aggregation should simultaneously satisfy: \textit{sparse} and \textit{informative}. To obtain the sparse representation with the spatial redundancy reduced, we propose a novel Compressed Global Context Aggregator~(CGCA) to convert the spatial feature maps containing  visual tokens into frequency domain and selectively preserve the low frequency information for global context reasoning. 

\begin{figure*}[htb]
	\begin{center}
		
		\includegraphics[width=0.83\linewidth]{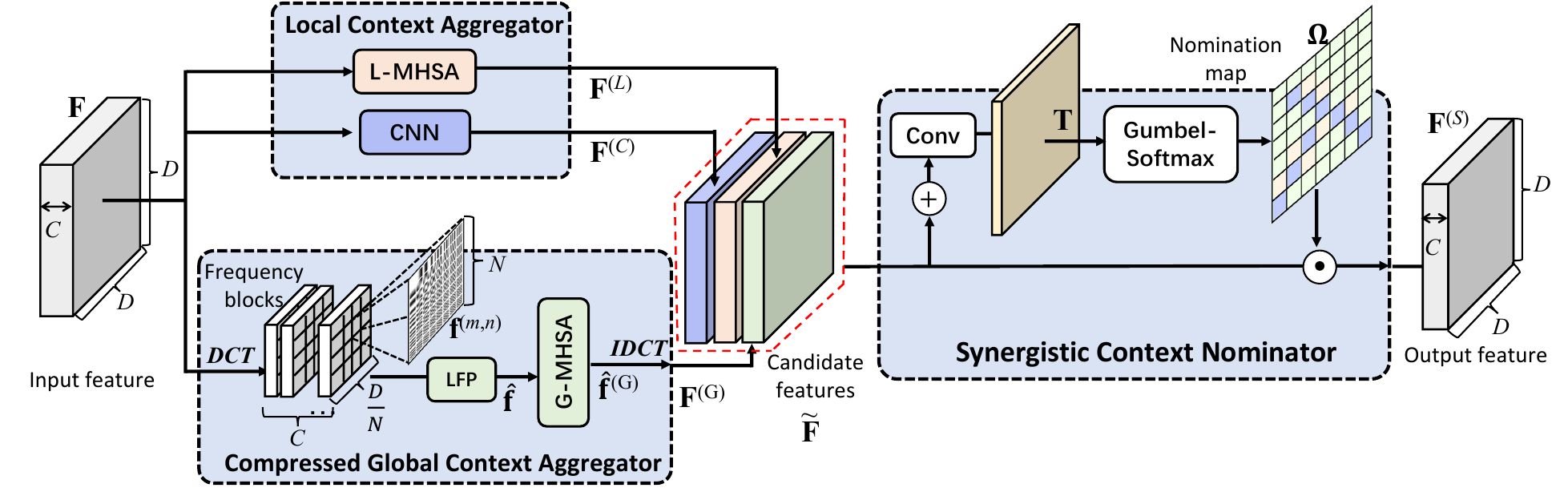}

	\end{center}
	\vspace{-2.5mm}
	\caption{Illustration of detailed Synergistic NomMer layer. Best viewed in color.} 
	\label{fig:nommer}
	\vspace{-2mm}
\end{figure*}

Unlike many previous methods, such as OctConv~\cite{chen2019drop}, conducting down-sampling on features, we aim at seeking the best trade-off between redundancy reduction and useful information preservation. The idea behind our method is mostly inspired by the spirit of JPEG codec~\cite{wallace1992jpeg} leveraging the discrete cosine transform~(DCT) to separate spatial frequencies from image. Specifically, as shown by ``Compressed Global Context Aggregator'' branch in Fig.~\ref{fig:nommer}, the input feature $\mathbf{F} \in \mathbb{R}^{D \times D \times C}$ is partitioned into blocks of $N \times N$ size  $ \{\mathbf{F}^{(m,n)},m = 1,2,...,\frac{D}{N},n = 1,2,...,\frac{D}{N}\},  \mathbf{F}^{(m,n)} \in \mathbb{R}^{N \times N \times C}$, where each channel of $\mathbf{F}^{(m,n)}$ is applied by 2D-DCT to obtain a serious of corresponding frequency blocks $\{\mathbf{f}^{(m,n)},m = 1,2,...,\frac{D}{N},n = 1,2,...,\frac{D}{N}\}, \mathbf{f}^{(m,n)} \in \mathbb{R}^{N \times N \times C}$, which is denoted as:
\begin{equation}
	\small
\begin{aligned}
			f(i, j)&=\sum_{u=0}^{N-1} \sum_{v=0}^{N-1} c(u) c(v)\\ 
		&\cdot F_{u, v} \cos \left[\frac{(i+0.5) \pi}{N} u\right] \cos \left[\frac{(j+0.5) \pi}{N} v\right],\\		
\end{aligned}
\end{equation}

\begin{equation}
	\small
	\begin{aligned}
		c(\lambda)&= \begin{cases}\sqrt{\frac{1}{N}}, & \lambda=0 \\ \sqrt{\frac{2}{N}}, & \lambda \neq 0\end{cases}, \label{eq:norm}
	\end{aligned}
\end{equation}
where $F_{u, v}$ represent the pixel with index $(u,v)$ in $\mathbf{F}^{(m,n)}$ while the $i$ and $j$ are indexes of horizontal and vertical spatial frequencies in frequency block $\mathbf{f}^{(m,n)}$. $c(\cdot)$ is the normalization scale factor ensuring the orthogonality.

Afterwards, the redundancy reduction is achieved by the low-frequency perceiver~(LFP) module. In detail, LFP first drops the high-frequencies of each frequency block in proportion $\alpha$, which is set to 0.5 in default: 
\begin{align}
	\small
\hat{\mathbf{f}}^{(m,n)} &=\{\mathbf{f}^{(m,n)}(i,j)\}, i,j \in \{1,2,...,l\},\\
l &= \lfloor\alpha N\rfloor.
\end{align}
then the low frequency map  is obtained by flattening each $\hat{\mathbf{f}}^{(m,n)}$ into a vector in $l^2 \cdot C$ dimension. To further extract useful frequencies while reducing dimensions, a convolutional layer with $1 \times 1 \times C$ kernel is applied to obtain the compressed frequency map $\hat{\mathbf{f}} \in \mathbb{R}^{\frac{D}{N} \times \frac{D}{N} \times C}$. As the redundancy is only reduced in the frequency domain while the spatial information still preserved, it is feasible to perform global context aggregation by using global multi-head self-attention~(G-MHSA): 
\begin{align}
	\small
	\hat{\mathbf{f}^{(G)}} = \textit{Conv}(\textit{G-MHSA}( \hat{\mathbf{f}})),
\end{align}
where $\hat{\mathbf{f}^{(G)}} \in \mathbb{R}^{\frac{D}{N} \times \frac{D}{N} \times N^2 \cdot C}$ and ``\textit{Conv}($\cdot$)'' is a convolutional layer with $1 \times 1 \times N^2 \cdot C$ kernel size. Then,  $\hat{\mathbf{f}^{(G)}}$ is reshaped to the tensor with shape ${\frac{D}{N} \times \frac{D}{N} \times N \times N \times C}$, where each block has the same shape~($N \times N \times C$) with $\mathbf{F}^{(m,n)}$. To project the frequency maps with compressed global context back to the spatial domain, the 2D-IDCT~(Inverse DCT) is conducted within each channel of each block $\hat{\mathbf{f}}^{(G)}_{(m,n)}$ according to:
\begin{equation}
	\small
\begin{aligned}
	 		F^{(G)}_{u, v} &=\sum_{u=0}^{N-1} \sum_{v=0}^{N-1} c(u) c(v)\\ 
	 &\cdot  f^{(G)}(i, j) \cos \left[\frac{(i+0.5) \pi}{N} u\right] \cos \left[\frac{(j+0.5) \pi}{N} v\right],\\
\end{aligned}
\end{equation}
where $F^{(G)}_{u, v}$ represent the pixel with index $(u,v)$ in a restored spatial feature block while the $f^{(G)}(i, j)$  are spatial frequencies in one frequency block of $\hat{\mathbf{f}^{(G)}}$. $c(\cdot)$ is the normalization scale factor given in Eqn.~\eqref{eq:norm}. Finally, spatial feature with compressed global context  $\mathbf{F}^{(G)}$ is yielded.

\subsubsection{Local Context Aggregator}
\vspace{-1mm}
Local Context Aggregator~(LCA) plays as the role of aggregating local context with more foveal vision-like details serving as the complementary visual information, which is expected to collaborate with compressed global context. To reach this goal, as illustrated in Fig.~\ref{fig:nommer}, we adopt two kinds of LCAs in this work, \textit{i.e.}, Local-MHSA~(L-MHSA)       
and CNN. As suggested by many prevalent researches~\cite{peng2021conformer,zhang2021aggregating}, CNN introduced in the ViT model can provide the inductive bias which is helpful for model convergence. Essentially, the CNN is the content-independent aggregator while the MHSA is the content-dependent one. The combination of CNN and MHSA can be deemed as a trade-off between leveraging pre-defined inductive bias or learning it from data. However, most of previous methods adopting parallel~\cite{peng2021conformer} or subsequent~\cite{zhang2021aggregating} combination manner can only have single acquisition w.r.t. the inductive bias within one layer. Contrastively, a more elegant way is to determine the inductive bias usage according to the specific region, which motivates our design of LCA providing both CNN feature $\mathbf{F}^{(C)} \in \mathbb{R}^{D \times D \times C}$ and L-MHSA feature~$\mathbf{F}^{(L)} \in \mathbb{R}^{D \times D \times C}$ for dynamic nomination. More concretely, we inherit the window-based self-attention~(W-MSA) with $M$ window size from SwinT~\cite{liu2021swin} as our L-MHSA. On the other hand, we adopt ``Bottleneck''~\cite{he2016deep} with structure \{Conv$_{1\times1}$, Conv$_{3\times3}$, Conv$_{1\times1}$\} as the CNN aggregator.

\vspace{-2mm}
\subsubsection{Synergistic Context Nominator}
\vspace{-1mm}
Once the compressed global and local context features obtained, they are treated as the candidate features~$\tilde{\mathbf{F}} = \{\mathbf{F}^{(L)}, \mathbf{F}^{(C)}, \mathbf{F}^{(G)}\} \in \mathbb{R}^{3 \times D \times D \times C}$ (highlighted by red dotted boundary in Fig.~\ref{fig:nommer}) for the proposed Synergistic Context Nominator~(SCN). The aim of SCN is to pick up the most valuable context with synergy for each spatial location, where the detailed process is described in Fig.~\ref{fig:nommer}. The candidate features are first fused by element-wise addition operation and passed to a convolutional layer with $1\times1\times3$ kernel to obtain a tensor $\mathbf{T}$ of shape $D \times D \times 3$, where each channels in the last axis describes the probability of nominating either one type context feature from three candidates. To further acquire the nomination map, we can perform the element-wise hard sampling on the tensor $\mathbf{T}$:
\begin{align}
	\small
\omega_{i,j} = \textit{argmax}(\tau_{i,j,1} , \tau_{i,j,2}, \tau_{i,j,3}),
\end{align}
 where $\tau_{i,j,c}$ is the $(i,j)$-th element in $c$-th channel of $\mathbf{T}$, $\omega_{i,j}$ is the channel index of nominated context feature at $(i, j)$ spatial location.  Correspondingly, the nomination map $\mathbf{\Omega} \in \mathbb{R}^{D \times D \times 3}$ can be determined. For example, if the L-MHSA context feature is nominated at position $(i, j)$, the corresponding nomination vector $\mathbf{\Omega}_{i,j}$ is $[1, 0, 0]$.
 
 One challenge lying in the proposed SCN module is that the hard sampling process is not differentiable whereas the weights in SCN need to be updated during training. To tackle this issue, we introduce a reparameterization method, termed Gumbel Softmax trick~\cite{jang2016categorical}, which allows the gradients to back-propagate through the discrete sampling process. In the end, the feature map $\mathbf{F}^{(S)} \in \mathbb{R}^{D \times D \times C} $ with nominated synergistic context is output by masking the nomination map on the candidate feature set:
 \begin{align}
 	\small
 	\mathbf{f}^{(S)}_{i,j} = \sum\nolimits_{p=1}^{3}  \mathbf{\Omega}_{i,j} \tilde{\mathbf{f}}_{p,i,j},
 \end{align}
  where $\mathbf{f}^{(S)}_{i,j}$ is the ($i,j$)-th feature vector of $\mathbf{F}^{(S)}$ and $p$ is the type index of context feature at location ($i,j$). 

\subsection{NomMer Variants}
To fully explore the potential of our proposed NomMer architecture with different configurations, we build several variants of it, \textit{i.e.}, NomMer-T, NomMer-S and NomMer-B, which refer to tiny, small and base model separately. The detailed configurations are given in supplementary material.

\section{Experiments}
\subsection{Image Classification on ImageNet-1K}
\paragraph{Experimental setting.}
We compare our proposed NomMer against several baselines on ImageNet-1K~\cite{deng2009imagenet}. For a fair comparison, we follow the experimental settings in~\cite{touvron2021training}. Concretely, all our models are pre-trained for 300 epochs with the input size of 224 $\times$ 224. The initial learning rate and batch size are set to $10^{-3}$ and 1,024, respectively. 
For optimization, AdamW~\cite{loshchilov2017decoupled} optimizer with a cosine learning rate scheduler is used. The weight decay is set to 0.05 and the maximal gradient norm is clipped to 5.0.
We also inherit the data augmentation and regularization techniques from~\cite{touvron2021training}. The stochastic depth drop rates are set to 0.1, 0.3 and 0.5 for tiny, small and base models individually. 
When reporting the results of 384 $\times$ 384 input, we fine-tune the models with a total batch size of 512 for 30 epochs. The learning rate and weight decay are $10^{-5}$ and $10^{-8}$.

\vspace{-4mm}
\paragraph{Performance.}
In Tab.~\ref{Comparison_Classification}, we compare our NomMer with state-of-the-art CNN and Transformer architectures on image classification task. The results show that the proposed NomMer consistently outperforms other approaches with similar model size and computational budgets. Compared with CNN-based RegNetY~\cite{radosavovic2020designing}, our models achieve improvements ranging from 1.6\% to 2.6\% under three configurations with input image in size of $224^2$. Compared with the ViT models, the proposed NomMer also witnesses the superior performance. More specifically, our NomMer-B with 73M parameters achieves a 84.5\% ImageNet Top-1 accuracy, surpassing DeiT-B~\cite{touvron2021training}, Swin-B~\cite{liu2021swin} and Conformer-B~\cite{peng2021conformer} by 2.7\%, 1.2\% and 0.4\%, respectively. Besides, the lightweight version~(NomMer-T) also achieves the best performance. When finetuned on the $384^2$ images, the similar trend is also observed. In addition, we further verify the effectiveness of our proposed method pre-trained on the larger ImageNet-21K dataset~(see supplementary material).
\begin{table}[htb]
	\small
	\setlength{\tabcolsep}{3.9mm}
	\centering
	\begin{tabular}{l|cc|c}
		\toprule[1.5pt]
		\multicolumn{4}{c}{\textbf{ImageNet-1K $\mathbf{224^{2}}$ trained models}} \\
		Method &  \tabincell{c}{\#param.\\ (M)} & \tabincell{c}{FLOPs\\(G)}   & \tabincell{c}{Top-1 \\(\%)} \\
		\hline
		RegNetY-4G~\cite{radosavovic2020designing} & 21 & 4.0  &80.0  \\
		RegNetY-8G~\cite{radosavovic2020designing} & 39 & 8.0 & 81.7 \\
		RegNetY-16G~\cite{radosavovic2020designing} & 84 & 16.0 & 82.9 \\
		\hline
		NFNet-F0~\cite{brock2021high} & 72 & 12.4  &83.6  \\
		DeiT-S~\cite{touvron2021training} & 22 & 4.6 & 79.8 \\
		DeiT-B~\cite{touvron2021training} & 86 & 17.5 & 81.8 \\
		\hline
		PVT-S~\cite{wang2021pyramid} & 25 & 3.8 & 79.8 \\
		PVT-M~\cite{wang2021pyramid} & 44 & 6.7 & 81.2 \\
		PVT-L~\cite{wang2021pyramid} & 61 & 9.8 & 81.7 \\
		\hline
		Swin-T~\cite{liu2021swin} & 29 & 4.5 & 81.3 \\
		Swin-S~\cite{liu2021swin} & 50 & 8.7 & 83.0 \\
		Swin-B~\cite{liu2021swin} & 88 & 15.4 & 83.3 \\
		
		\hline
		T2T-ViT$_{t}$-14~\cite{yuan2021tokens} & 22 & 6.1 & 81.7 \\
		T2T-ViT$_{t}$-19~\cite{yuan2021tokens} & 39 & 9.8 & 82.2 \\
		T2T-ViT$_{t}$-24~\cite{yuan2021tokens} & 64 & 15.0 & 82.6 \\
		\hline
		LG-T~\cite{li2021local} & 33 & 4.8 & 82.1 \\
		LG-S~\cite{li2021local} & 61 & 9.4 & 83.3 \\
		\hline
		Focal-T~\cite{yang2021focal} & 29 & 4.9 & 82.2 \\
		Focal-S~\cite{yang2021focal} & 51 & 9.1 & 83.5 \\
		Focal-B~\cite{yang2021focal} & 90 & 16.0 & 83.8 \\
		\hline
		Conformer-T~\cite{peng2021conformer} & 24 & 5.2 & 81.3 \\
		Conformer-S~\cite{peng2021conformer} & 38 & 10.6 & 83.4 \\
		Conformer-B~\cite{peng2021conformer} & 83 & 23.3 & 84.1 \\
		\hline
		NesT-T~\cite{zhang2021aggregating} &  17 & 5.8 & 81.5 \\
		NesT-S~\cite{zhang2021aggregating} &  38 & 10.4 & 83.3 \\
		NesT-B~\cite{zhang2021aggregating} &  68 &  17.9 & 83.8 \\
		\hline
		CvT-13~\cite{wu2021cvt} & 20 & 4.5 & 81.6  \\
		CvT-21~\cite{wu2021cvt} & 32 & 7.1 & 82.5 \\
		\hline
		CaiT-S~\cite{touvron2021going} &  68 & 13.9 & 84.0 \\
		
		\hline
		NomMer-T & 22 & 5.4  &\textbf{82.6} \\
		NomMer-S & 42 & 10.1 &\textbf{83.7} \\
		NomMer-B & 73 & 17.6 & \textbf{84.5} \\
		
		\midrule[1.5pt]
		\multicolumn{4}{c}{\textbf{ImageNet-1K $\mathbf{384^{2}}$ finetuned models}} \\
		ViT-B/16~\cite{dosovitskiy2020image} & 86 & 49.3 & 77.9 \\
		\hline
		DeiT-B~\cite{touvron2021training} & 86 & 55.4 & 83.1 \\
		\hline
		Swin-B~\cite{liu2021swin} & 88 & 47.0 & 84.2 \\
		\hline
		T2T-ViT$_{t}$-14~\cite{yuan2021tokens} & 22 & 17.1 & 83.3 \\
		\hline
		CvT-13~\cite{wu2021cvt} & 20 & 16.3 & 83.0  \\
		CvT-21~\cite{wu2021cvt} & 32 & 24.9 & 83.3 \\
		\hline
		CaiT-S~\cite{touvron2021going} &  68 & 48.0 & 85.4 \\
		\hline
		NomMer-T & 22 & 17.2 & \textbf{83.9} \\
		NomMer-S & 42 & 33.1 &\textbf{84.6} \\			
		NomMer-B & 73 & 56.2 & 84.9 \\
		\bottomrule[1.5pt]
	\end{tabular}
	\caption{Comparison of different backbones on ImageNet-1K classification.}
	\label{Comparison_Classification}
\end{table}
\subsection{Object Detection on COCO}
\paragraph{Experimental setting.}
To verify NomMer's versatility, we benchmark it on object detection with COCO 2017~\cite{lin2014microsoft}. The models pretrained on ImageNet-1K~\cite{deng2009imagenet} are used for initializing the backbone of Cascade Mask R-CNN~\cite{cai2018cascade} framework. Similar to SwinT~\cite{liu2021swin}, we follow 3$\times$ schedule training  with 36 epochs for a fair comparison. 
During training, multi-scale training strategy is employed to randomly resize image's shorter side to the range of [480, 800]. And we use AdamW~\cite{loshchilov2017decoupled} for optimization with initial learning rate $10^{-4}$ and weight decay 0.05.
 In the similar spirit, 0.1, 0.3 and 0.5 stochastic depth drop rates are set to regularize the training for tiny, small and base models.

\vspace{-3mm}
\paragraph{Performance.}
 The box and mask mAPs on COCO validation set are summarized in Tab.~\ref{Comparison_Detection}, from which we can see that NomMer significantly boosts the $AP^{b}$ and $AP^{m}$.  In details, the box's mAP and mask's mAP of NomMer-B are 0.8\% and 0.6\% higher than that of the strong baseline Swin-B~\cite{liu2021swin}, which demonstrates the importance of global representations for high level tasks in our method. When evaluating our tiny model, it surpasses the second best method Focal-T~\cite{yang2021focal} by 0.3\%, which also suggests that NomMer can also perform well on the prediction tasks with fewer parameters. To further investigate the versatility of the proposed model when it works in different detection frameworks, we also conduct a series of experiments to compare NomMer with other SOTAs. {A more detailed description is given in supplementary material.}
 
\begin{table}[htb]
	\small
	\setlength{\tabcolsep}{0.2mm}
		\begin{tabular}{l|cc|ccc|ccc}
			\toprule[1.5pt]
			Method & \tabincell{c}{\#param.\\ (M)} & \tabincell{c}{FLOPs\\(G)} & \tabincell{c}{$AP^{b}$\\ (\%)} & \tabincell{c}{$AP_{50}^{b}$ \\(\%)} & \tabincell{c}{$AP_{75}^{b}$ \\(\%)}& \tabincell{c}{$AP^{m}$\\ (\%)} & \tabincell{c}{$AP_{50}^{m}$ \\(\%)} & \tabincell{c}{$AP_{75}^{m}$ \\(\%)}\\
			\hline
			Res50~\cite{he2016deep} & 82 & 739 & 46.3 & 64.3 & 50.5 & 40.1 & 61.7 & 43.4 \\
			X101-32~\cite{xie2017aggregated} & 101 & 819 & 48.1 & 66.5 & 52.4 & 41.6 & 63.9 & 45.2 \\
			X101-64~\cite{xie2017aggregated} & 140 & 972 & 48.3 & 66.4 & 52.3 & 41.7 & 64.0 & 45.1 \\
			\hline
			Swin-T~\cite{liu2021swin} & 86&745&50.5&69.3&54.9&43.7&66.6&47.1 \\
			Swin-S~\cite{liu2021swin} & 107 &838 &51.8&70.4&56.3&44.7&67.9&\textbf{48.5} \\
			Swin-B~\cite{liu2021swin} & 145&982&51.9&70.9&56.5&45.0&68.4&48.7 \\

			\hline
			Focal-T~\cite{yang2021focal} & 87 &770 & 51.5&70.6&55.9&-&-&- \\
			\hline

			NomMer-T & 80&755 &\textbf{51.8} &	\textbf{70.8}&	\textbf{56.0}&	\textbf{44.7}&	\textbf{67.6}&	\textbf{48.1} \\
			NomMer-S & 99&851& \textbf{52.4}	&\textbf{71.5}&	\textbf{56.8}&	\textbf{45.1}&	\textbf{68.8}&	\textbf{48.5} \\
			NomMer-B & 130&1006& \textbf{52.7}	&\textbf{71.6}&	\textbf{57.2}	& \textbf{45.6}	&\textbf{68.9}	&\textbf{49.3} \\
			\bottomrule[1.5pt]
		\end{tabular}
		\centering
		\caption{Results on COCO object detection and instance segmentation with Cascade Mask R-CNN.}
		\label{Comparison_Detection}
\end{table}

\vspace{-4mm}
\subsection{Semantic Segmentation on ADE20K}
\paragraph{Experimental setting.}
For another dense prediction task, Semantic Segmentation, we further evaluate our model on the ADE20K~\cite{zhou2017scene} dataset. 
Specifically, our NomMer serves as the backbone of UperNet~\cite{xiao2018unified} which is a prevalent segmentation method. Unless explicitly specified, we use a standard recipe by setting the image size to 512 $\times$ 512 and train the models for 160k iterations with batch size 16.

\vspace{-3mm}
\paragraph{Performance.}
The results of Upernet with various backbones on the ADE20K~\cite{zhou2017scene} dataset are reported in the Tab.~\ref{Comparison_Segmentation}, where both single-scale and multi-scale evaluation results are included.
Obviously, our method significantly outperforms previous state-of-the-arts under different configurations. Under single-scale setting, our NomMer separately achieves 50.0\%, 48.7\%, and 46.1\% mIoU with base, small and tiny model configurations, which are 1.0\%, 0.7\%, 0.3\% higher than the strong baseline Focal~\cite{yang2021focal} counterparts. And we can also observe the consistent performance improvement under the multi-scale evaluation. Conclusively, our NomMer can steadily improve the performance of various visual recognition tasks owing to the synergistic context nomination.

\begin{table}[htb]
	\small
	\setlength{\tabcolsep}{3.3mm}
	\begin{tabular}{l|cc|cc}
		\toprule[1.5pt]
		Method &\tabincell{c}{\#param.\\ (M)} & \tabincell{c}{FLOPs\\(G)} & \tabincell{c}{mIoU\\(\%)}  & \tabincell{c}{+MS\\(\%)} \\
		\hline
		Res101~\cite{he2016deep} & 86 & 1029 & 44.9 & -\\
		\hline
		Swin-T~\cite{liu2021swin} &60&945& 44.5 &45.8\\
		Swin-S~\cite{liu2021swin} &81&1038& 47.6 &49.5\\
		Swin-B~\cite{liu2021swin} &121&1188& 48.1 &49.7\\
		\hline
		Focal-T~\cite{yang2021focal} &62&998& 45.8 &47.0\\
		Focal-S~\cite{yang2021focal} &85&1130& 48.0 &50.0\\
		Focal-B~\cite{yang2021focal} &126&1354& 49.0 &50.5\\
		\hline
		LG-T~\cite{li2021local} & 64& 957&45.3& -\\
		\hline
		NomMer-T &54&954& \textbf{46.1} & \textbf{47.3} \\
		NomMer-S &73&1056& \textbf{48.7} & \textbf{50.4} \\
		NomMer-B &107&1220& \textbf{50.0} & \textbf{51.0} \\
		\bottomrule[1.5pt]
	\end{tabular}
	\caption{Performance comparison of different backbones with UperNet framework on the ADE20K segmentation task.}
	\label{Comparison_Segmentation}
	\centering
\end{table}

\begin{table*}[htb]
	\small
	\setlength{\tabcolsep}{2.2mm}
	\begin{tabular}{l|cc|cc|c|c|cc|c}
		\toprule[1.5pt]
		\multirow{2}{*}{Method} & \multicolumn{2}{c|}{Local} & \multicolumn{2}{c|}{Global} & Nominator & ImageNet & \multicolumn{2}{c|}{COCO} & ADE20K \\
		& L-MHSA & CNN & Pool & DCT & Gumbel & Top-1(\%) & $AP^{b}$ & $AP^{m}$& mIoU(\%) \\
		\hline
		$\text{NomMer-T}^*$  w/o Global\&CNN & \cmark& \xmark& \xmark& \xmark&  \xmark& 81.4& 50.4 & 43.5 & 44.7 \\
		$\text{NomMer-T}^*$  w/o Global & \cmark& \cmark& \xmark& \xmark&  \xmark& 81.7& 50.6 & 43.8 & 44.8 \\
		$\text{NomMer-T}^*$  w/o Gumbel\&DCT & \cmark& \cmark& \cmark& \xmark&  \xmark& 82.0& 50.9 & 44.0 & 45.1 \\
		$\text{NomMer-T}^*$  w/o Gumbel & \cmark& \cmark& \xmark& \cmark&  \xmark& 82.2& 51.2 & 44.3 & 45.4 \\
		\hline
		\textbf{NomMer-T} & \cmark& \cmark& \xmark& \cmark& \cmark& \textbf{82.6}& \textbf{51.8} & \textbf{44.7} & \textbf{46.1} \\
		\bottomrule[1.5pt]
	\end{tabular}
\vspace{-1.5mm}
	\caption{Ablation study of NomMer on three benchmarks based on the NomMer-T architecture.}
	\label{Ablation}
	\centering
\end{table*}


\begin{figure*}[htb]
	\begin{center}
		\includegraphics[width=1\linewidth]{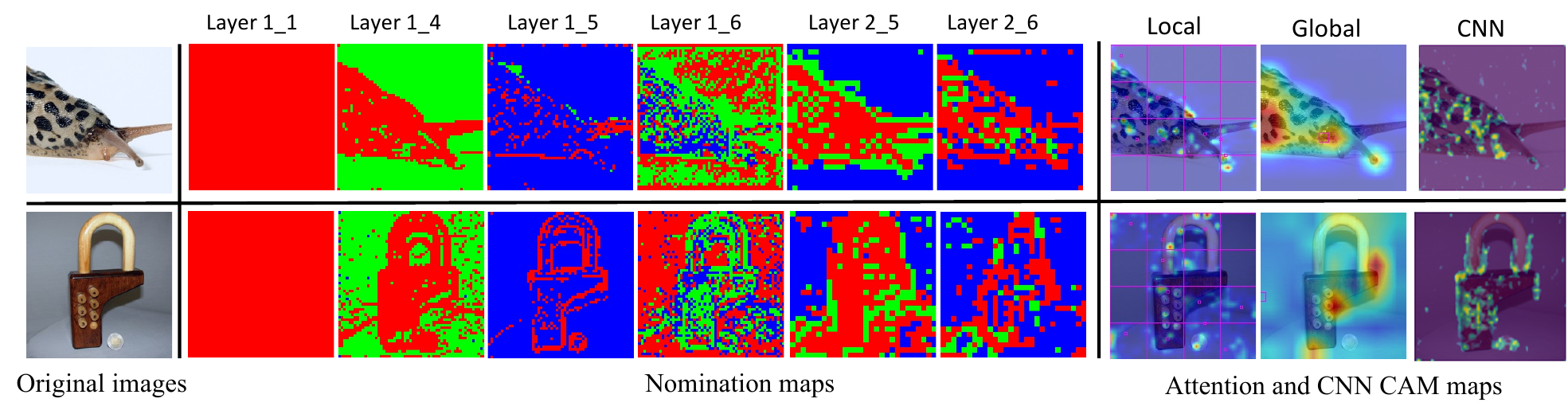}
	\end{center}
	\vspace{-6.5mm}
	\caption{Visualization of nomination maps, attention maps and CNN CAM~\cite{wang2020score} maps of nominated features from NomMer-B on classification task. \textbf{Red}: CNN context $\mathbf{F}^{(C)}$. \textbf{Green}: Local context $\mathbf{F}^{(L)}$. \textbf{Blue}: Compressed global context $\mathbf{F}^{(G)}$. ``Layer $B\_L$'' stands for that map is from the $L$-th NomMer layer of  NomMer blocks at the $B$-th stage. The pink hollow boxes in attention maps represent the locations of nominated local or global context features. Best viewed in color and zoom in.} 
	\label{fig:mommap}
	\vspace{-0.5mm}
\end{figure*}

\subsection{Ablation Study}
To better investigate the effectiveness of different aspects in our proposed S-NomMer layer, we conduct extensive ablation studies on both classification and downstream tasks of which the results are summarized in Tab.~\ref{Ablation}. 
\vspace{-3mm}
\paragraph{Effect of Local Context Aggregator.} We validate the effectiveness of the LCA combing both L-MHSA and CNN. Compared with the performance of ``-w/o Global\&CNN'' only equipped with L-MHSA, the additional CNN~(``-w/o Global'') consistently increases 0.3\% accuracy on ImageNet~\cite{deng2009imagenet}, 0.2\% box's mAP and 0.3\% mask's mAP on COCO~\cite{lin2014microsoft}, and 0.1\% mIoU on ADE20K~\cite{zhou2017scene}. This confirms the benefit of inductive bias provided by CNN and also shows the combination can capture the salient fine-grained features through synergy.
\vspace{-3mm}
\paragraph{Effect of Compressed Global Context Aggregator.} 
As is shown in Tab.~\ref{Ablation}, we find the performance of all tasks can be consistently improved by integrating the global context into local ones, even if the global context is aggregated from simple max-pooling features. When the aggregation is performed in the frequency domain through our DCT-based CGCA, the average accuracy of each task is further boosted, which indicates our learnable CGCA can well strike the trade-off between redundancy reduction and useful information preservation for the visual recognition tasks.

\vspace{-3mm}
\paragraph{Effect of Synergistic Context Nominator.}
To verify the capability of another core component SCN in our architecture, we compare the performance of our NomMer with and without the SCN equipped on each task. As demonstrated in Tab.~\ref{Ablation}, our model can obtain the best results by adopting nominator, with at least 0.4\% improvement on different tasks than the non-nominator version in which various types of contexts are directly fused through element-wise addition.
These results further prove the effectiveness of the synergistic context learned by SCN.

  \begin{figure}[htb]
	\begin{center}
		
		\includegraphics[width=0.95\linewidth]{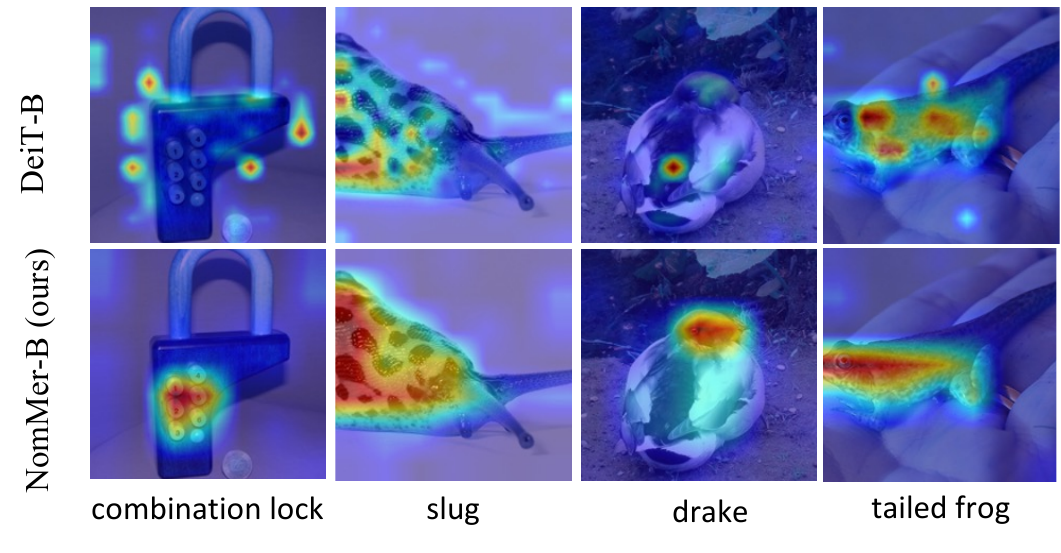}
	\end{center}
	\vspace{-6.5mm}
	\caption{Class activation attention maps on classification task. Best viewed in color.} 
	\label{fig:clsmap}
	\vspace{-1mm}
\end{figure}

\vspace{-1.5mm}
\subsection{Qualitative Analysis}\label{qual_ana}
To further investigate the working pattern of our proposed NomMer, in Fig.~\ref{fig:mommap}, we visualize the synergistic nomination maps from intermediate layers of S-NomMer block from base model on classification task. We surprisingly observe that nomination maps exhibit several intriguing properties. In low-level nomination maps~(``Layer~1\_1'') the CNN context features are always predominant, which is consistent with the conclusion given in research~\cite{xiao2021early} , that early convolutions help transformers see better. Along with the model going deeper, nomination maps present various synergistic patterns of context in different layers. Specifically, the local context features aggregated by CNN and L-MHSA contribute most in ``Layer 1\_4'', while the adjacent ``Layer 1\_5'' focuses on the global context accompanied with local CNN context. These phenomena demonstrate that our model can successfully obtain synergistic context with redundancy reduced, where the synergy not only happens between local and global context, but also cross layers. 

Moreover, one can also observe that the global context features are inclined to be nominated in the smooth regions, \textit{e.g.},~``Layer 1\_5'' with less color and texture variations. We attribute this behavior to the missing of salient details in smooth regions where model needs to refer more information from larger scope, which is illustrated by the global attention maps in Fig.~\ref{fig:mommap}~ from ``Layer 2\_5''.  Comparatively, there are more fine-grained texture or patterns involved in the local context aggregated by CNN and L-MHSA, which can be demonstrated by the local attention maps and CNN CAM maps~(from ``Layer 2\_5'') obtained by algorithm~\cite{wang2020score}. 

Thanks to the nomination mechanism, NomMer-B can capture more discriminative features from synergistic context than other ViTs, such as DeiT~\cite{touvron2021training} only exploiting global context. By visualizing attention maps of class activation based on method~\cite{chefer2021transformer}, in Fig.~\ref{fig:clsmap}, we find that the attentions of NomMer-B often concentrate on the exclusive features of object, such as the keyboard of `` combination lock'' and the dot pattern of ``slug'' while DeiT-B presents more unstable activation maps.

\vspace{-1.5mm}
\section{Conclusion}
\vspace{-1.5mm}
In this work, we introduced a novel vision transformer architecture solving visual recognition tasks by nominating synergistic context. Extensive experiments on image classification and dense prediction tasks demonstrated its superiority over state-of-the-arts. The visualization of the nomination maps learned by our method can also provide an empirical guidance for the design of architecture, which will be further explored in our future work.

\begin{appendices}

\section{Preliminary: Multi-Head Self-Attention}
As introduced in the main text, Multi-Head Self-Attention~(MHSA)~\cite{vaswani2017attention} is the core component of the ViT model, on which we build our local context aggregator~(LCA) and compressed global context aggregator~(CGCA) in the Synergistic Context Nominator~(SCN), as well as the G-NomMer layer in our method. As the preliminary knowledge, we briefly review it in the following. Given queries $\mathbf{Q}$, keys $\mathbf{K}$ and values $\mathbf{V}$, Multi-Head Attention~(MHA) is formulated as:
\begin{align*}
	\small
	\centering
	&\textit{MultiHead}(\mathbf{Q,K,V}) = \textit{Concat}(\mathbf{H}_1,\mathbf{H}_2,...,\mathbf{H}_h)\mathbf{W^*},\\
	& \mathbf{H}_i = \textit{Attention}(\mathbf{QW}_i^Q,\mathbf{KW}_i^K,\mathbf{VW}_i^V ),  i \in \{1,2,..., h\},\\
	&\textit{Attention}(\mathbf{Q,K,V}) = \textit{softmax}(\frac{\mathbf{Q}\mathbf{K}^\top}{\sqrt{d_k}})\mathbf{V},\label{eq:att}
\end{align*}
where $h$ is the head number and $d_k$ is the key dimension. $\mathbf{W}_i^Q \in \mathbb{R}^{d_m \times d_k},\mathbf{W}_i^K \in \mathbb{R}^{d_m \times d_k},\mathbf{W}_i^V \in \mathbb{R}^{d_m \times d_v}$ and $\mathbf{W}_i^* \in \mathbb{R}^{hd_v \times d_m}$ are corresponding projection matrices. In particular, the MHA becomes MHSA on the condition that $\mathbf{X}=\mathbf{Q}=\mathbf{K}=\mathbf{V}$, where $\mathbf{X}$ denotes the input. In the G-MHSA~(Fig.~\ref{fig:nommer} in main text) and  G-NomMer layer~(Fig.~\ref{fig:arch}(b) in main text) of our proposed architecture, the relations of all tokens within global scope are captured by MHSA while the local dependencies of the tokens falling inside the window are built by L-MHSA~(Fig.~\ref{fig:nommer} in main text).   

\section{More Implementation Details}
The proposed NomMer architecture is implemented by Pytorch~\cite{paszke2019pytorch} and all experiments are conducted on a workstation with 32 NVIDIA A100-80 GB GPUs. The detailed configurations of a series of NomMer variants and the core pseudo code of Synergistic NomMer block are elaborated as follows.
\subsection{Detailed Configurations of NomMer Variants}
The detailed configurations of three variants of NomMer are given in Tab.~\ref{tab:architecture}, including NomMer-T, NomMer-S and NomMer-B, which refer to tiny, small and base model separately. In the ``S-NomMer'', the parameters of three types of candidate context aggregators~(from top to bottom are ``CNN'', ``L-MHSA'', ``G-MHSA'') are separated by dotted lines.

\subsection{Pseudo Code of Synergistic NomMer Block}

\begin{minipage}{.95\linewidth}
	\begin{algorithm}[H]
		\KwInput{$\mathbf{F}$}
		\KwOutput{$\mathbf{F}^{(S)}$}
		
		\SetKwFunction{FCGCA}{{CGCA}}
		\SetKwFunction{FSCN}{{SCN}}
		\SetKwFunction{FMain}{{S-NomMer}}
		
		\tcc{Compressed Global Context Aggregator.}
		\Fn{\FCGCA{$\mathbf{F}$}} {
			\tcc{Discrete Cosine Transform.}
			$\mathbf{f} ~\leftarrow~ \textit{DCT}(\mathbf{F})$ 
			
			\tcc{Low-Frequency Perceiver.}
			$\mathbf{\hat{f}} ~\leftarrow~ \textit{LFP}(\mathbf{f})$ 
			
			$\mathbf{\hat{f}}^{(G)} ~\leftarrow~ \textit{Conv}(\textit{G-MHSA}(\mathbf{\hat{f}}))$ 
			
			\tcc{Inverse Discrete Cosine Transform.}
			$\mathbf{F}^{(G)} ~\leftarrow~ \textit{IDCT}(\mathbf{\hat{f}}^{(G)})$ 
			
			\KwRet $\mathbf{F}^{(G)}$
		}
		
		\tcc{Synergistic Context Nominator.}
		\Fn{\FSCN{$\mathbf{{F}}^{(L)}, \mathbf{{F}}^{(C)}, \mathbf{{F}}^{(G)}$}}{
			$\mathbf{\Omega} ~\leftarrow~ \textit{Conv}(\mathbf{{F}}^{(L)} + \mathbf{{F}}^{(C)} + \mathbf{{F}}^{(G)})$ 
			
			$\mathbf{{F}}^{(S)} ~\leftarrow~ \textit{Nominate}(\mathbf{\Omega}, \mathbf{{F}}^{(L)}, \mathbf{{F}}^{(C)}, \mathbf{{F}}^{(G)})$ 
			
			\KwRet $\mathbf{{F}}^{(S)}$
		}
		\tcc{Synergistic NomMer.}
		\Fn{\FMain}{
			$\mathbf{{F}}^{(L)} ~ \leftarrow~ \textit{L-MHSA}(\mathbf{F})$
			
			$\mathbf{{F}}^{(C)} ~\leftarrow~ \textit{CNN}(\mathbf{F})$
			
			$\mathbf{{F}}^{(G)} ~\leftarrow~ \textit{CGCA}(\mathbf{F})$
			
			$\mathbf{{F}}^{(S)} ~\leftarrow~ \textit{SCN}(\mathbf{{F}}^{(L)}, \mathbf{{F}}^{(C)}, \mathbf{{F}}^{(G)})$
			
			$\mathbf{{F}}^{(S)} ~\leftarrow~ \mathbf{{F}}^{(S)} + \textit{FFN}(\mathbf{{F}}^{(S)})$
			
			\KwRet $\mathbf{{F}}^{(S)}$
		}
		\caption{S-NomMer pseudo code.}
		\label{alg}
	\end{algorithm}
\end{minipage}
\section{Image Classification on ImageNet-21K}

\paragraph{Experimental setting.}
We further pre-train NomMer on the larger ImageNet-21K dataset, which contains 14.2M images and 21K classes. During pre-training stage, we employ the AdamW~\cite{loshchilov2017decoupled} optimizer with a weight decay of $10^{-2}$, an initial learning rate of $10^{-2}$, and a batch size of 4,096 for 90 epochs with the input size of $224^2$. In ImageNet-1K fine-tuning, we train the models for 30 epochs with a batch size of
1,024, a constant learning rate of $10^{-5}$, and a weight decay of $10^{-8}$ on $224^2$ and $384^2$ input.

\begin{table}[htb]
	\small
	\setlength{\tabcolsep}{5mm}
	\centering
	\begin{tabular}{l|cc|c}
		\toprule[1.5pt]
		\multicolumn{4}{c}{\textbf{ImageNet-21K $\mathbf{224^{2}}$ finetuned models}} \\
		Method & \tabincell{c}{\#param.\\ (M)} & \tabincell{c}{FLOPs\\(G)} & \tabincell{c}{Top-1 \\(\%)} \\
		\hline
		Swin-B~\cite{liu2021swin} & 88 & 15.4 & 85.2 \\
		\hline
		NomMer-B  &73 & 17.6 & \textbf{85.5} \\	
		
		\midrule[1.5pt]
		\multicolumn{4}{c}{\textbf{ImageNet-21K $\mathbf{384^{2}}$ finetuned models}} \\
		Method  & \tabincell{c}{\#param.\\ (M)} & \tabincell{c}{FLOPs\\(G)} & \tabincell{c}{Top-1 \\(\%)} \\
		\hline
		R-101x3~\cite{kolesnikov2020big}& 388 & 204.6 & 84.4 \\
		ViT-B/16~\cite{dosovitskiy2020image}& 86 & 55.4 & 84.0 \\
		ViL-B~\cite{zhang2021multi}& 56 & 43.7 & 86.2 \\
		Swin-B~\cite{liu2021swin}& 88 & 47.1 & 86.4 \\
		\hline
		NomMer-B &73 & 56.2 & \textbf{86.6} \\	
		\bottomrule[1.5pt]
	\end{tabular}
	\caption{Comparison of different backbones on ImageNet-21K classification.}
	\label{app:Comparison_Classification}
\end{table}

\paragraph{Performance.}
The results of pre-training on ImageNet-21K are summarized in Tab.~\ref{app:Comparison_Classification},
from which we can see that our NomMer-B obtains 85.5\% and 86.6\% top-1 accuracy under $224^2$ and $384^2$ input size setting. Compared with the second best method, Swin~\cite{liu2021swin}, our method can outperform it by at least 0.2\%.

\section{Object Detection on COCO with Various Frameworks}
\paragraph{Experimental setting.}
To further verify the effectiveness of our proposed NomMer when working in different detection frame works, we conduct extensive experiments by training four typical object detectors on COCO dataset, including Cascade Mask R-CNN~\cite{cai2018cascade}, ATSS~\cite{zhang2020bridging}, RepPointsV2~\cite{yang2019reppoints} and Sparse R-CNN~\cite{sun2021sparse}.
For a fair comparison, we utilize the same experimental settings for all four detectors. More concretely, all the models are training with 3$\times$ schedule, multi-scale training, AdamW~\cite{loshchilov2017decoupled} optimizer, initial learning rate of 0.0001, weight decay of 0.05, and batch size of 16.
\paragraph{Performance.}
The box mAPs on COCO are reported in Tab.~\ref{app:Comparison_Detection}. As one can see, the NomMer-T witnesses the substantive improvements on the performance of different detectors, which demonstrates our NomMer architecture is a versatile backbone for various object detection approaches.

\begin{table}[htb]
	\small
	\setlength{\tabcolsep}{0.3mm}
	\begin{tabular}{c|l|cc|ccc}
		\toprule[1.5pt]
		Method & Backbone & \tabincell{c}{\#param.\\ (M)} & \tabincell{c}{FLOPs\\(G)} & \tabincell{c}{$AP^{b}$\\ (\%)} & \tabincell{c}{$AP_{50}^{b}$ \\(\%)} & \tabincell{c}{$AP_{75}^{b}$ \\(\%)}\\
		\hline
		\multirow{4}{*}{ \tabincell{c}{Cascade Mask \\  R-CNN~\cite{cai2018cascade}}} &Res50~\cite{he2016deep} & 82 & 739 & 46.3 & 64.3 & 50.5 \\
		&Swin-T~\cite{liu2021swin} & 86&745&50.5&69.3&54.9  \\
		&Focal-T~\cite{yang2021focal} & 87 &770 & 51.5&70.6&55.9\\
		&NomMer-T & 80&755 &\textbf{51.8} &	\textbf{70.8}&	\textbf{56.0} \\
		\hline
		\multirow{4}{*}{ATSS~\cite{zhang2020bridging}}&Res50~\cite{he2016deep} & 32 & 205 & 43.5 & 61.9 & 47.0 \\
		&Swin-T~\cite{liu2021swin} & 36&212&47.2&66.5&51.3  \\
		&Focal-T~\cite{yang2021focal} & 37 &239 & 49.5&\textbf{68.8}&53.9\\
		&NomMer-T & 30&237 &\textbf{49.8} &	{68.6}&	\textbf{54.0} \\
		\hline
		\multirow{4}{*}{RepPointsV2~\cite{yang2019reppoints}} &Res50~\cite{he2016deep} & 43 & 431 & 46.5 & 64.6 & 50.3 \\
		&Swin-T~\cite{liu2021swin} & 44&437&50.0&68.5&54.2 \\
		&Focal-T~\cite{yang2021focal} & 45 &491 & 51.2&70.4&54.9\\
		&NomMer-T & 41&486 &\textbf{51.6} &	\textbf{70.7}&	\textbf{55.1} \\
		\hline
		\multirow{4}{*}{ \tabincell{c}{Sparse  \\ R-CNN~\cite{sun2021sparse}}} &Res50~\cite{he2016deep} & 106& 166 & 44.5 & 63.4& 48.2 \\
		&Swin-T~\cite{liu2021swin} & 110&172&47.9&67.3&52.3  \\
		&Focal-T~\cite{yang2021focal} & 111 &196 & 49.0&69.1&53.2\\
		&NomMer-T & 104&195 &\textbf{49.5} &	\textbf{69.3}&	\textbf{53.7} \\
		\bottomrule[1.5pt]
	\end{tabular}
	\centering
	\caption{Results on COCO object detection across different object detection methods.}
	\label{app:Comparison_Detection}
\end{table}

\begin{table*}[htb]
	\setlength{\tabcolsep}{0.65mm}
	\small
	\begin{tabular}{c|c|c|c|c|c}
		\toprule
		& \tabincell{c}{Output \\ Size} & \tabincell{c}{Layer \\ Name} &NomMer-T & NomMer-S & NomMer-B \\
		\hline
		& 56 * 56 & \tabincell{c}{Patch\\ Embedding} & dim 96, conv 4*4& dim 96, conv 4*4& dim 128, 4*4 \\
		\hline
		\multirow{3}{*}{stage1} & 56*56 & \tabincell{c}{S-NomMer \\ Block} &$\left[ \tabincell{c}{$\left[\tabincell{c}{$\textrm{dim~96, ~conv~1*1,}$ \\ $\textrm{conv~3*3, ~conv~1*1}$}\right]$ \\ \dotline[2pt]{3.2cm} \\ $\left [\textrm{dim~96,~head~2,~wsize~7}\right]$ \\ \dotline[2pt]{3.2cm} \\ $\left [\textrm{dim~96,~head~2,~ksize~8} \right]$}\right]\times 2$ & $\left[ \tabincell{c}{$\left[\tabincell{c}{$\textrm{dim~96, ~conv~1*1,}$ \\ $\textrm{conv~3*3, ~conv~1*1}$}\right]$ \\ \dotline[2pt]{3.2cm} \\ $\left [\textrm{dim~96,~head~2,~wsize~7}\right]$ \\ \dotline[2pt]{3.2cm} \\ $\left [\textrm{dim~96,~head~2,~ksize~8} \right]$}\right]\times 6$ &$\left[ \tabincell{c}{$\left[\tabincell{c}{$\textrm{dim~128, ~conv~1*1,}$ \\ $\textrm{conv~3*3, ~conv~1*1}$}\right]$ \\ \dotline[2pt]{3.2cm} \\ $\left [\textrm{dim~128,~head~2,~wsize~7}\right]$ \\ \dotline[2pt]{3.2cm} \\ $\left [\textrm{dim~128,~head~2,~ksize~8} \right]$}\right]\times 6$  \\
		\cline{2-6} 
		& 28*28 & \tabincell{c}{Reduction \\ Module} & $\textrm{dim\ 192, \ conv 3*3, \ pool \ /2}$  & $\textrm{dim\ 192, \ conv 3*3, \ pool \ /2}$  &$\textrm{dim\ 256, \ conv 3*3, \ pool \ /2}$  \\
		\hline
		\multirow{3}{*}{stage2} & 28*28 & \tabincell{c}{S-NomMer \\ Block} & $\left[ \tabincell{c}{$\left[\tabincell{c}{$\textrm{dim~192, ~conv~1*1,}$ \\ $\textrm{conv~3*3, ~conv~1*1}$}\right]$ \\ \dotline[2pt]{3.2cm} \\ $\left [\textrm{dim~192,~head~2,~wsize~7}\right]$ \\ \dotline[2pt]{3.2cm} \\ $\left [\textrm{dim~192,~head~2,~ksize~4} \right]$}\right]\times 2$ & $\left[ \tabincell{c}{$\left[\tabincell{c}{$\textrm{dim~192, ~conv~1*1,}$ \\ $\textrm{conv~3*3, ~conv~1*1}$}\right]$ \\ \dotline[2pt]{3.2cm} \\ $\left [\textrm{dim~192,~head~2,~wsize~7}\right]$ \\ \dotline[2pt]{3.2cm} \\ $\left [\textrm{dim~192,~head~2,~ksize~4} \right]$}\right]\times 6$ & $\left[ \tabincell{c}{$\left[\tabincell{c}{$\textrm{dim~256, ~conv~1*1,}$ \\ $\textrm{conv~3*3, ~conv~1*1}$}\right]$ \\ \dotline[2pt]{3.2cm} \\ $\left [\textrm{dim~256,~head~2,~wsize~7}\right]$ \\ \dotline[2pt]{3.2cm} \\ $\left [\textrm{dim~256,~head~2,~ksize~4} \right]$}\right]\times 6$   \\
		\cline{2-6}
		& 14*14 &\tabincell{c}{Reduction \\ Module} & $\textrm{dim\ 384, \ conv 3*3, \ pool \ /2}$  & $\textrm{dim\ 384, \ conv 3*3, \ pool \ /2}$  &$\textrm{dim\ 512, \ conv 3*3, \ pool \ /2}$ \\
		\hline
		\multirow{2}{*}{stage3} & 14*14&\tabincell{c}{G-NomMer \\ Block} & $\left [ \textrm{dim\ 384},\ \textrm{head\ 4} \right] \times 8 $  & $\left [ \textrm{dim\ 384},\ \textrm{head\ 4} \right] \times 16 $  & $\left [ \textrm{dim\ 512},\ \textrm{head\ 4} \right] \times 16 $   \\
		\cline{2-6}
		&  7*7& \tabincell{c}{Reduction \\ Module} & $\textrm{dim\ 768, \ conv 3*3, \ pool \ /2}$  & $\textrm{dim\ 768, \ conv 3*3, \ pool \ /2}$  &$\textrm{dim\ 1024, \ conv 3*3, \ pool \ /2}$  \\
		\hline
		stage4& 7*7 & \tabincell{c}{G-NomMer \\ Block} & $\left [ \textrm{dim\ 768},\ \textrm{head\ 8} \right] \times 2 $ & $\left [ \textrm{dim\ 768},\ \textrm{head\ 8} \right] \times 4 $ & 	$\left [ \textrm{dim\ 1024},\ \textrm{head\ 8} \right] \times 4 $    \\
		\bottomrule
	\end{tabular}
	\caption{Detailed architecture specifications of NomMer.}
	\label{tab:architecture}
	\centering
\end{table*}

\section{More Visual Interpretability}
\subsection{Nomination Maps on Different Tasks}
\begin{figure*}[htp]
	\centering
	\includegraphics[width=0.9\linewidth]{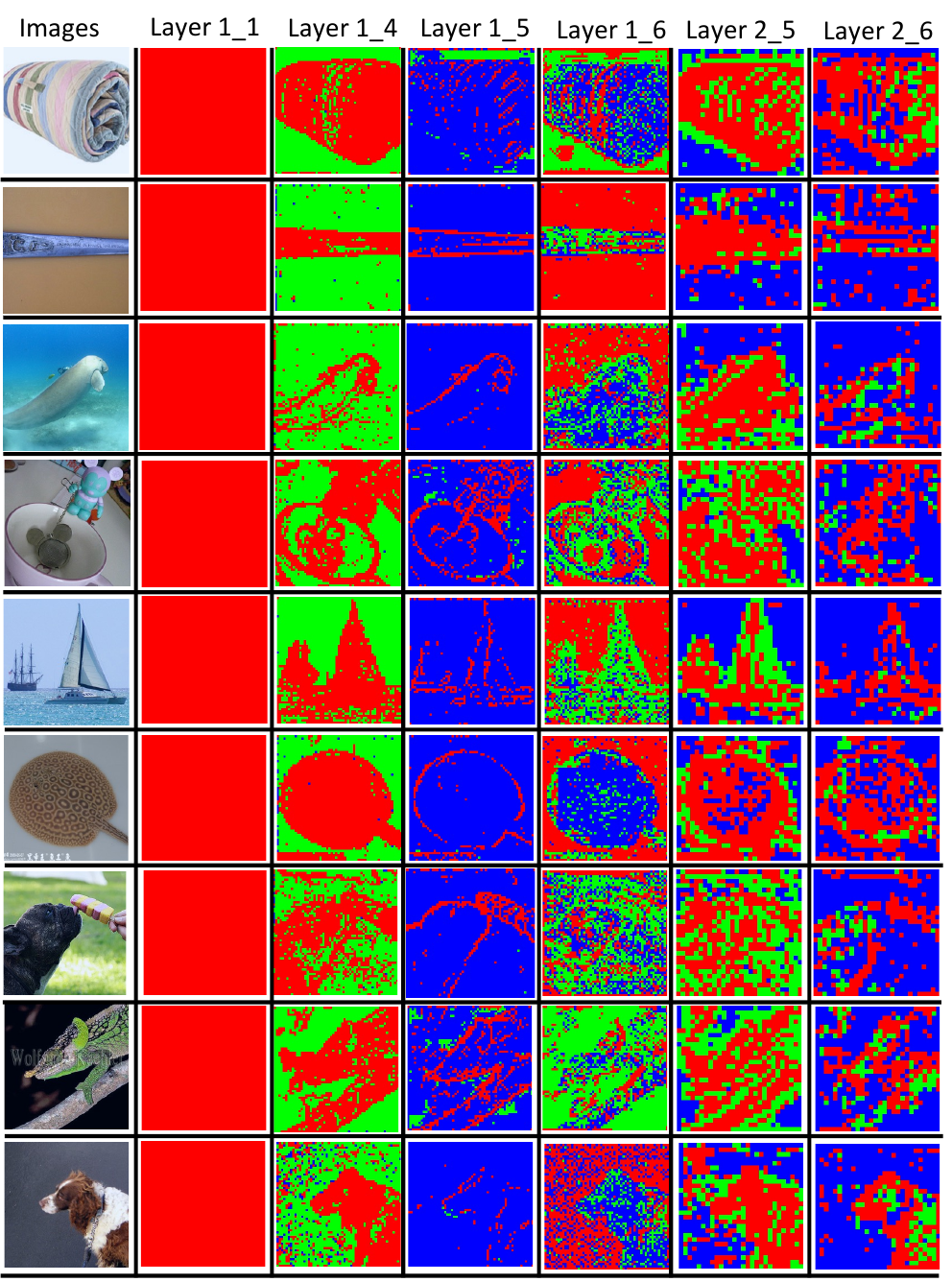}
	\caption{Nomination Maps from NomMer-B on Image Classification task. \textbf{Red}: CNN context $\mathbf{F}^{(C)}$. \textbf{Green}: Local context $\mathbf{F}^{(L)}$. \textbf{Blue}: Compressed global context $\mathbf{F}^{(G)}$. ``Layer $B\_L$'' stands for that map is from the $L$-th NomMer layer of NomMer blocks at the $B$-th stage. Best viewed in color.}
	\label{fig:nommap}
\end{figure*}

For the image classification task, we visualize more nomination maps in Fig.~\ref{fig:nommap} of this appendix, which also follows the similar patterns described in the Sec.~\ref{qual_ana} of main text.

\begin{figure*}[htp]
	\centering
	\includegraphics[width=0.9\linewidth]{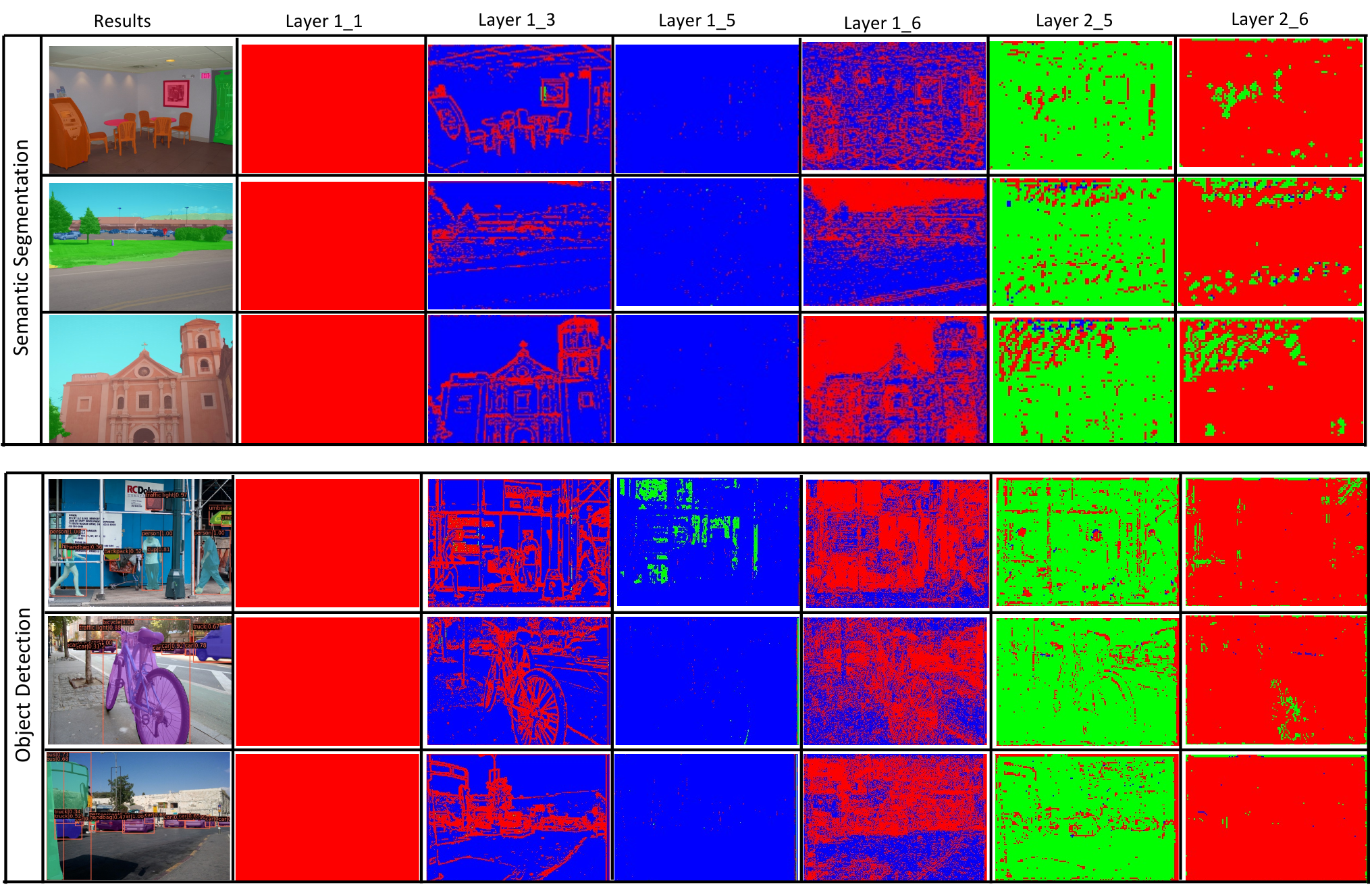}
	\caption{Nomination Maps from NomMer-B on dense prediction tasks including semantic segmentation and object detection. \textbf{Red}: CNN context $\mathbf{F}^{(C)}$. \textbf{Green}: Local context $\mathbf{F}^{(L)}$. \textbf{Blue}: Compressed global context $\mathbf{F}^{(G)}$. ``Layer $B\_L$'' stands for that map is from the $L$-th NomMer layer of NomMer block at the $B$-th stage. Best viewed in color.}
	\label{fig:nommap2}
\end{figure*}

Additionally, in Fig.~\ref{fig:nommap2} of this appendix, we also visualize the nomination maps from intermediate layers of S-NomMer blocks when semantic segmentation and object detection frameworks adopting NomMer-B as backbones~(corresponding to the results in Tab.~\ref{Comparison_Detection} and Tab.~\ref{Comparison_Segmentation} of main text). Compared to the maps on classification task, the nomination maps of dense prediction tasks exhibit different context synergy patterns. Although the CNN context features are also predominant in low-level nomination maps~(``Layer~1\_1''), the context aggregated by L-MHSA is hardly observed in the first stage layers~(``Layer~1$\_\sim$''). Instead, most regions of ``Layer~1\_3'' maps mainly pick up global context accompanied with CNN context focusing on the salient object details, such as outlines of wheels or buildings. The ``Layer~1\_5'' maps further witness the domination of global context. 
We attribute these phenomena to the larger spatial size of early-stage blocks providing sufficiently precise spatial information, which is indispensable to the dense prediction tasks. This explanation can also be confirmed by the Non-local Neural Networks~\cite{wang2018non}, where the most significant improvement on performance is achieved by inserting the SA-based non-local block in the early stage. In contrast, the ``non-local'' behavior in our method is implemented in a more graceful way with object details taken into account. Compared with the ``Layer~1\_3'' maps, the ``Layer~1\_6'' maps are inclined to emphasize CNN contextual features containing details at a more fine-grained level. In the layers of the second stage~(``Layer~2$\_\sim$''), the local CNN and L-MHSA contextual features are mostly nominated, where the maps from adjacent layers become complementary, which is different from the synergistic context pattern on classification task.

In summary, Fig.~\ref{fig:nommap} and Fig.~\ref{fig:nommap2} vividly show that our NomMer can successfully modulate the synergy behavior of different types of context in terms of specific tasks.

\subsection{Representation Structure of NomMer}
\begin{figure*}[htp]
	\centering
	\includegraphics[width=1\linewidth]{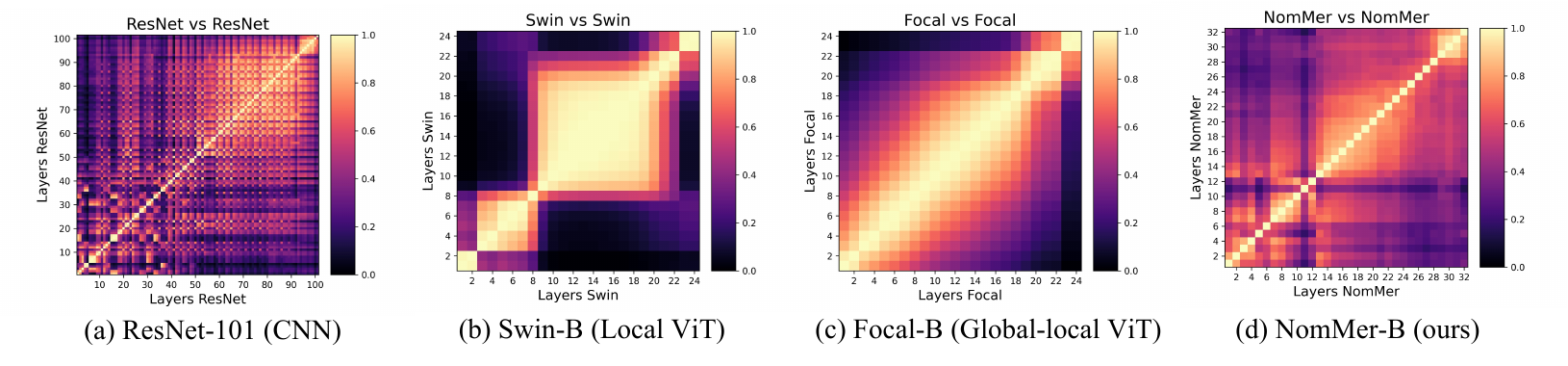}
	\caption{CKA similarities between all pairs of layers across different model architectures trained on ImageNet-1k. The results are shown as heatmaps in which the horizontal and vertical axes indexing the layers from input to output. Best viewed in color and zoom in.}
	\label{fig:cka}
\end{figure*}
As aforementioned, the synergy behavior can appear either within or across layers, which has been demonstrated by the Fig.~\ref{fig:nommap} and Fig.~\ref{fig:nommap2}. To further study the representation structure learned in our NomMer, we plot the Centered Kernel Alignment~(CKA) similarities~\cite{kornblith2019similarity} between all pairs of layers across different model architectures in Fig.~\ref{fig:cka}. Given the representations of two layers $\mathbf{X} \in \mathbb{R}^{m \times c_1}$ and $\mathbf{Y} \in \mathbb{R}^{m \times c_2}$, mathematically, 
\begin{align*}
	\textit{CKA}(\mathbf{K}, \mathbf{L})&=\frac{\textit{HSIC}(\mathbf{K}, \mathbf{L})}{\sqrt{\textit{HSIC}(\mathbf{K}, \mathbf{K}) \textit{HSIC}(\mathbf{L}, \mathbf{L})}},\\
	\mathbf{K}& = \mathbf{X}\mathbf{X}^\top, \mathbf{L} = \mathbf{Y}\mathbf{Y}^\top, 
\end{align*} 
where $\textit{HSIC}$ is the Hilbert-Schmidt Independence Criterion which measures the similarity of centered similarity matrices: 
\begin{align*}
	\textit{HSIC} (\mathbf{K}, \mathbf{L})&= \frac{\textit{vec}(\mathbf{K}^\prime)\cdot \textit{vec}(\mathbf{L}^\prime)}{(m-1)^2},\\
	\mathbf{K}^\prime  &= \mathbf{H}\mathbf{K}\mathbf{H}, \mathbf{L}^\prime  = \mathbf{H}\mathbf{L}\mathbf{H}, 
\end{align*}
and $\mathbf{H}=\mathbf{I}_{n}-\frac{1}{n} \mathbf{11}^\top$ is the centering matrix.

From Fig.~\ref{fig:cka}~(b), we can observe that the local self attention-based ViT, Swin-B~\cite{liu2021swin}, presents the similarity structure with clear block-like patterns, and the similarity scores are always high within a ``block'' which contains several adjacent layers while almost become zero outside the block. By introducing global contextual information, Focal-B~\cite{yang2021focal}~( Fig.~\ref{fig:cka}~(c)) also exhibits block-like similarity structure but with more smooth edges, indicating that there still have representation similarity between layers with larger interval. Compared with ViT models, in the canonical CNN architecture, ResNet-101~\cite{he2016deep}~( Fig.~\ref{fig:cka}~(a)), the representation within lower layers present more difference while the similarities tend to be larger between higher layers. Moreover, the similarity scores between lower and higher layers are small.

By comparison, unlike the ViT models with heuristic-based design in terms of exploiting local or global-local context, the representation structure of our NomMer-B has somewhat similarity with that of ResNet-101 while the ``block'' patterns in ViT are also preserved. This more sophisticated structure can be attributed to the ``dynamic nomination'' of NomMer, which effectively integrate the contributions of local and global context.      

\subsection{More Qualitative Analysis on NomMer }
\begin{figure}[htp]
	
	\includegraphics[width=1.02\linewidth]{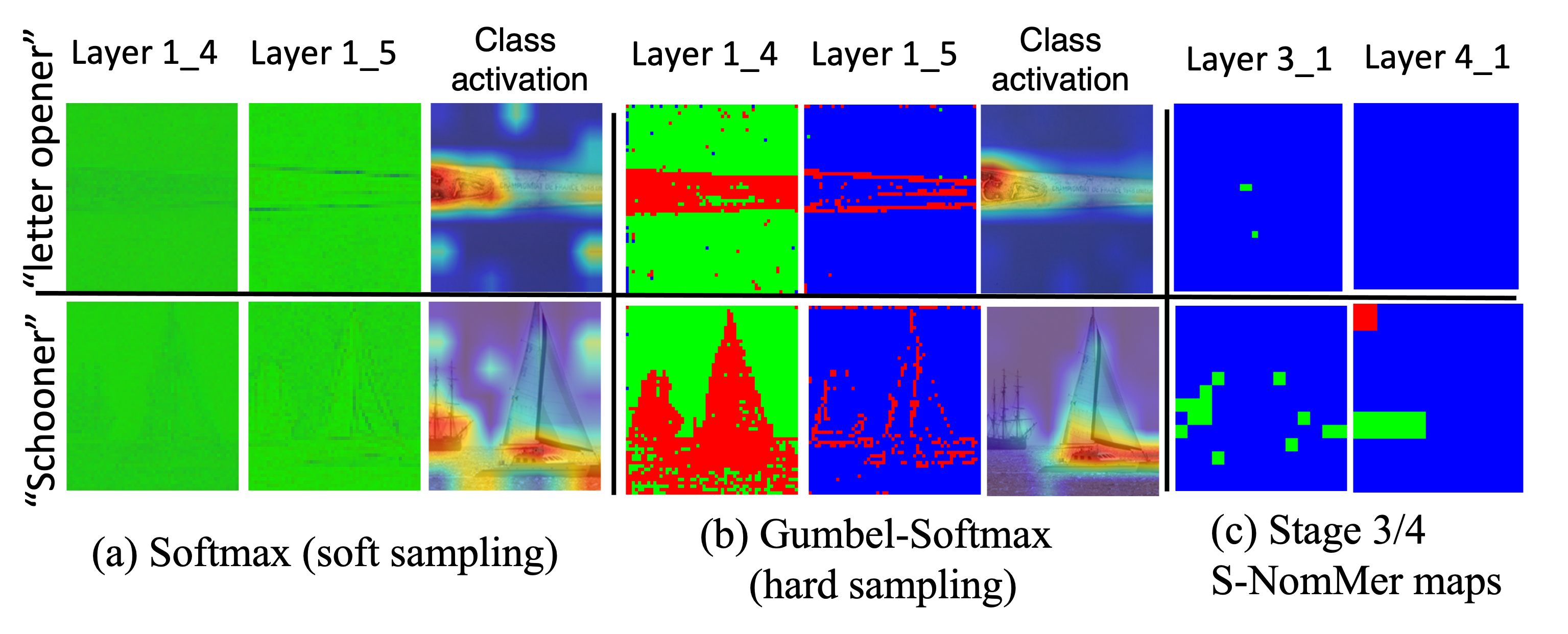}
	
	\small
	\caption{Nomination maps (soft sampling~(a) vs. hard sampling~(b)) and class activation attention maps from NomMer-B on image classification task. (c) Nomination maps at stage 3 and 4.} 
	\label{fig:hard_soft}
\end{figure}

This part will further illustrate more evidence on indispensable design of nominator and G-NomMer block in our method. 
\paragraph{Hard sampling vs. soft sampling.}
By replacing the Gumbel-softmax in SCN of NomMer with canonical softmax, we find that, in contrast to the hard version~(Fig.~\ref{fig:hard_soft}(b)), the local-attention features (green) present dominant in the learned maps of soft version~(Fig.~\ref{fig:hard_soft}(a)). Correspondingly, the classification activation maps of soft version become more unstable than hard version. As a result, the top1 accuracies of different NomMer variants on image classification task all drop round 0.4 on image classification task. We attribute it to the redundancy not well reduced by soft sampling, where local and global features could have negative impact on each other.   
\paragraph{G-NomMer block.}
As shown in Fig.~\ref{fig:hard_soft}(c), if the S-NomMer blocks were applied to all stages, the global SA features (blue) would be dominant at both stage 3 (Layer3\_1) and stage 4 (Layer4\_1). It is probably because the global context and local context could become homogeneous when feature size become small at higher-level stage 3 and 4, as claimed in main text. Therefore, we adopt the G-NomMer block only equipped with canonical global SA.
{\small
	\bibliographystyle{ieee_fullname}
	\bibliography{cvpr}
}
\end{appendices}
\end{document}